\documentclass[11pt,a4paper]{article}
\usepackage{emnlp-ijcnlp-2019}
\usepackage{times}
\usepackage{latexsym}
\usepackage{url}
\usepackage{algorithm, algorithmic}
\usepackage{graphicx}
\usepackage{multirow}
\usepackage{subfig}
\usepackage{wrapfig}
\usepackage{array}
\usepackage{dblfloatfix}
\usepackage{url}
\usepackage{amsmath}
\newsavebox{\measurebox}
\newcommand{\mturkresult}{70.40}

\aclfinalcopy

\newcommand*{\sepfbox}[1]{%
  \begingroup
    \sbox0{\fbox{#1}}%
    \setlength{\fboxrule}{0pt}%
    \fbox{\unhbox0}%
  \endgroup
}

\title{The Trumpiest Trump? \\ Identifying a Subject's Most Characteristic Tweets}
\author{Charuta Pethe \\
  Department of Computer Science, \\ Stony Brook University, NY, USA \\
  {\tt cpethe@cs.stonybrook.edu} \\\And
  Steven Skiena \\
  Department of Computer Science, \\ Stony Brook University, NY, USA \\
  {\tt skiena@cs.stonybrook.edu} \\}

\date{}

\begin{document}
\maketitle

\begin{abstract}
The sequence of documents produced by any given author varies in style and content, but some documents are more typical or representative of the source than others.  We quantify the extent to which a given short text is characteristic of a specific person, using a dataset of tweets from fifteen celebrities.  Such analysis is useful for generating excerpts of high-volume Twitter profiles, and understanding how representativeness relates to tweet popularity. We first consider the related task of binary author detection (is \textit{x} the author of text \textit{T}?), and report a test accuracy of 90.37\% for the best of five approaches to this problem. We then use these models to compute characterization scores among all of an author's texts. A user study shows human evaluators agree with our characterization model for all 15 celebrities in our dataset, each with p-value $<$ 0.05. We use these classifiers to show surprisingly strong correlations between characterization scores and the popularity of the associated texts. Indeed, we demonstrate a statistically significant correlation between this score and tweet popularity (likes/replies/retweets) for 13 of the 15 celebrities in our study.
\end{abstract}

\section{Introduction}
\label{sec:intro}
Social media platforms, particularly microblogging services such as Twitter, have become increasingly popular \cite{statista} as a means to express thoughts and opinions. Twitter users emit tweets about a wide variety of topics, which vary in the extent to which they reflect a user's personality, brand and interests. This observation motivates the question we consider here, of how to quantify the degree to which tweets are characteristic of their author?

People who are familiar with a given author appear to be able to make such judgments confidently.
For example, consider the following pair of tweets written by US President Donald Trump, at the extreme sides of our characterization scores (0.9996 vs. 0.0013) for him:

\vspace{0.1in}
\setlength{\fboxsep}{0.5em}
\noindent\sepfbox{%
\parbox{0.93\linewidth}{%
\textbf{Tweet 1:} Thank you for joining us at the Lincoln Memorial tonight- a very special evening! Together, we are going to MAKE AMERICA GREAT AGAIN!
}%
}
\noindent\sepfbox{%
\parbox{0.93\linewidth}{%
\textbf{Tweet 2:} ``The bend in the road is not the end of the road unless you refuse to take the turn." - Anonymous
}%
}
\newline

Although both these tweets are from the same account, we assert that Tweet 1 sounds more characteristic of Donald Trump than Tweet 2. We might also guess that the first is more popular than second. Indeed, Tweet 1 received 155,000 likes as opposed to only 234 for Tweet 2.

Such an author characterization score has many possible applications. With the ability to identify the most/least characteristic tweets from a person, we can generate reduced excerpts for high-volume Twitter profiles.  Similarly, identifying the least characteristic tweets can highlight unusual content or suspicious activity.  A run of sufficiently unrepresentative tweets might be indicative that a hacker has taken control of a user's account.

But more fundamentally, our work provides the necessary tool to study the question of how ``characteristic-ness" or novelty are related to tweet popularity. Do tweets that are more characteristic of the user get more likes, replies and retweets?   Is such a relationship universal, or does it depend upon the personality or domain of the author?
Twitter users with a large follower base can employ our methods to understand how characteristic a new potential tweet sounds, and obtain an estimate of how popular it is likely to become.

To answer these questions, we formally define the problem of author representativeness testing, and model the task as a binary classification problem. Our primary contributions in this paper include:

\begin{itemize}

\item \textbf{Five approaches to authorship verification:} As a proxy for the question of representativeness testing (which has no convincing source of ground truth without extensive human annotation), we consider the task of distinguishing tweets written by a given author from others they did not write.  We compare five distinct computational approaches to such binary tweet classification (user vs. non-user). Our best model achieves a test accuracy of 90.37\% over a dataset of 15 Twitter celebrities. We use the best performing model to compute a score (the probability of authorship), which quantifies how characteristic of the user a given tweet is.

\item \textbf{Human evaluation study:} To verify that our results are in agreement with human judgment of how `characteristic' a tweet is, we ask human evaluators which of a pair of tweets sounds more characteristic of the given celebrity. The human evaluators are in agreement with our model \mturkresult\% of the time, significant above the $0.05$ level for each of our 15 celebrities.

\item \textbf{Correlation analysis for popularity:} Our characterization score exhibits strikingly high absolute correlation with popularity (likes, replies and retweets), despite the fact that tweet text is the only feature used to train the classifier which yields these scores.
\begin{figure}[H]
\includegraphics[width=\linewidth]{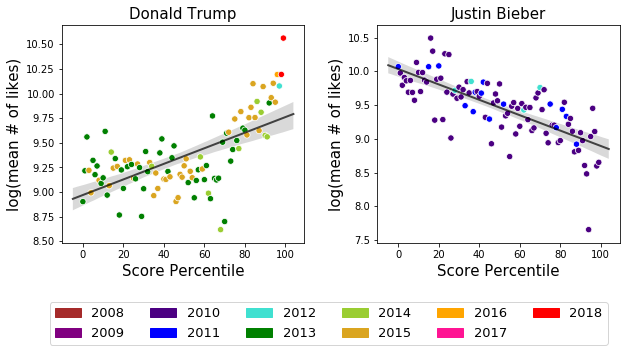}
\caption{Plot of log mean number of likes against tweet score percentile for Donald Trump and Justin Bieber.  Node color denotes the year for which the maximum number of tweets are present in each percentile bucket, demonstrating that this is not merely a temporal correlation.}
  \label{fig:trump_bieber_corr}
\end{figure}
For 13 of the 15 celebrities in our dataset, we observe a statistically significant correlation between characterization score and popularity. Figure \ref{fig:trump_bieber_corr} shows the relation between tweet score and tweet popularity for Donald Trump and Justin Bieber respectively. The figure shows that the sign of this association differs for various celebrities, reflecting whether their audience seeks novelty or reinforcement.

\item \textbf{Iterative sampling for class imbalance:}  Our task requires distinguishing a user's tweets (perhaps 1,000 positive training examples) from the sea of all other user's tweets (implying billions of possible negative training examples). We present an iterative sampling technique to exploit this class imbalance, which improves the test accuracy for negative examples by 2.62\%.

\end{itemize}

\section{Problem Formulation}
\label{sec:probform}
We formally define the author representativeness problem as follows:
\newline
\newline
\textbf{Input:}
 A Twitter author $U$ and the collection of their tweets, and
 a new tweet $T$.

\noindent
\textbf{Problem:}
Compute $\textrm{score}(T, U)$, the probability that $T$ was written by $U$. This score quantifies how characteristic of writer $U$, tweet $T$ is.

\subsection{Methodology}
\label{sec:methodology}
In order to obtain this representativeness score, we model our task as a classification problem, where we seek to distinguish tweets from $U$ against tweets from all other users.

By modeling this as a binary classification problem, it becomes possible to quantify how characteristic of a writer a tweet is, as a probability implied by its distance from the decision boundary. Thus, we obtain a characterization score between 0 and 1 for each tweet.

\noindent\textbf{Challenges:} In training a classifier to distinguish between user and non-user tweets, we should ideally have an equal amount of examples of both classes. User tweets are simply all the tweets from that user's Twitter account, and measure perhaps in the thousands. Indeed, the number of tweets per user per day is limited to 2400 per day by current Twitter policy (\href{https://help.twitter.com/en/rules-and-policies/twitter-limits}{https://help.twitter.com/en/rules-and-policies/twitter-limits}).
The negative examples consist of all tweets written by other Twitter users, a total of approximately 500 million per day (\href{https://business.twitter.com/}{https://business.twitter.com}).
Thus there is an extreme class imbalance between user and non-user tweets. Moreover, the nature of language used on Twitter does not conform to formal syntactic or semantic rules. The sentences tend to be highly unstructured, and the vocabulary is not restricted to a particular dictionary.

\subsection{Data}
\label{sec:data}

For the binary classification task described in Section \ref{sec:methodology}, we term tweets from $U$ as positive examples, and tweets from other users as negative examples.

\begin{itemize}
    \item \textbf{Positive examples:} We take tweets written by 15 celebrities from various domains, from 01-Jan-2008 to 01-Dec-2018, as positive examples. Properties of these Twitter celebrities are provided in Table \ref{tab:celebs}.
    \item \textbf{Negative examples:} We have collected 1\% of tweets from Twitter's daily feed using the Twitter API (\href{https://developer.twitter.com/en/docs.html}{https://developer.twitter.com/en/docs.html}) to use as negative examples.
\end{itemize}
\begin{table}[H]
\centering
\resizebox{\linewidth}{!}{%
\begin{tabular}{|c|cc|c|c|}
\hline
\multirow{2}{*}{\textbf{User}} & \multicolumn{2}{c|}{\textbf{Tweet count}} & \multirow{2}{*}{\textbf{Domain}} & \multirow{2}{*}{\textbf{\begin{tabular}[c]{@{}c@{}}Foll.\end{tabular}}} \\
 & \textbf{(Before)} & \textbf{(After)} &  &  \\ \hline
Amitabh Bachchan & 49437 & 10342 & Acting & 37.0 \\
Ariana Grande & 37738 & 13657 & Music & 62.3 \\
Barack Obama & 11350 & 6772 & Politics & 106.0 \\
Bill Gates & 2699 & 1754 & Business & 47.1 \\
Donald Trump & 35549 & 18295 & Politics & 59.9 \\
Ellen DeGeneres & 17317 & 9616 & TV & 77.6 \\
J K Rowling & 6037 & 2634 & Author & 14.7 \\
Jimmy Fallon & 10698 & 3596 & TV & 51.1 \\
Justin Bieber & 18044 & 5193 & Acting & 105.0 \\
Kevin Durant & 22532 & 4146 & Sports & 17.5 \\
Kim Kardashian & 24541 & 7943 & Modeling & 60.5 \\
Lady Gaga & 7239 & 3767 & Music & 78.5 \\
LeBron James & 5145 & 2102 & Sports & 42.6 \\
Narendra Modi & 17613 & 6672 & Politics & 47.0 \\
Oprah Winfrey & 11685 & 4588 & TV & 42.2 \\ \hline
\end{tabular}%

}
\caption{Twitter celebrities in our dataset, with tweet counts before and after filtering (Foll. denotes followers in millions)}
\label{tab:celebs}
\end{table}
\noindent
\textbf{Preprocessing and Filtering:}
We have preprocessed and filtered the data to remove tweets that are unrepresentative or too short for analysis. All text has been converted to lowercase, and stripped of punctuation marks and URLs. This is because our approaches are centered around word usage. However, in future models, punctuation may prove effective as a feature. Further, we restrict analysis to English language tweets containing no attached images. We select only tweets which are more than 10 words long, and contain at least 5 legitimate (dictionary) English words. We define an unedited transfer of an original tweet as a retweet, and remove these from our dataset. Since comments on retweets are written by the user themselves, we retain these in our dataset.

We note that celebrity Twitter accounts can be handled by PR agencies, in addition to the owner themselves. Because our aim is to characterize Twitter profiles as entities, we have not attempted to distinguish between user-written and agency-written tweets. However, this is an interesting direction for future research.

We use a train-test split of 70-30\% on the positive examples, and generate negative training and test sets of the same sizes for each user, by randomly sampling from the large set of negative examples.

\section{Related work}
\label{sec:relwork}

\subsection{Author identification and verification}
\label{sec:authid}
The challenge of author identification has a long history in NLP. PAN 2013 \cite{pan2013} introduced the question: ``Given a set of documents by the same author, is an additional (out-of-set) document also by that author?'' The corpus is comprised of text pieces from textbooks, newspaper articles, and fiction. Submissions to PAN 2014 \cite{pan2014} also model authorship verification as binary classification, by using non-author documents as negative examples. The best submission \cite{pan2013winner} in PAN 2013 uses the General Impostors (GI) method, which is a modification of the Impostors Method \cite{impostorsmethod}. The best submission \cite{slightlymodifiedimpostor} in PAN 2014 presents a modification of the GI method. These methods are based on the impostors framework \cite{giframework}.

\newcite{compressionveenman} used compression distance as a document representation, for authorship verification in PAN 2013. \newcite{worddist} present a global feature extraction approach and achieve state-of-the-art accuracy for the PAN 2014 corpus. The best submission \cite{rnnauthid} in PAN 2015 \cite{pan2015} uses a character-level RNN model for author identification, in which each author is represented as a sub-model, and the recurrent layer is shared by all sub-models. This is useful if the number of authors is fixed, and the problem is modeled as multi-class classification. \newcite{deeplearningauthid} also approach multi-class author identification, using deep learning for feature extraction, and \newcite{forensic} using hierarchical clustering.

\newcite{intrinsicauthorverification} propose an intrinsic profile-based verification method that uses latent semantic indexing (LSI), which is effective for longer texts. \newcite{authoneclass} and \newcite{limiteddata} explore methods for authorship verification for larger documents such as essays and novels. \newcite{emailauthid} and \newcite{email2} explore author identification for emails, and \newcite{taskguided} for scientific papers. \newcite{azarbonyad2015time} make use of temporal changes in word usage to identify authors of tweets and emails. \newcite{unigramsandbigrams},  \newcite{featureeval}, and \newcite{unstructured} evaluate the utility of various features for this task. \newcite{authidusingtextsampling} proposes text sampling to address the lack of text samples of undisputed authorship, to produce a desirable distribution over classes.

\newcite{koppel2009computational} compare methods for variants of the authorship attribution problem. \newcite{bhargava2013stylometric} apply stylometric analysis to tweets to determine the author. \newcite{lopez2015discriminative} propose a document representation capturing discriminative and subprofile-specific information of terms. \newcite{rocha2016authorship} review methods for authorship attribution for social media forensics. \newcite{peng2016bit} use bit-level n-grams for determining authorship for online news. \newcite{peng2016astroturfing} apply this method to detect astroturfing on social media. \newcite{theophilo2019needle} employ deep learning specifically for authorship attribution of short messages.

\subsection{Predicting tweet popularity}
\label{sec:predpop}
\newcite{suh2010want} leverages features such as URL, number of hashtags, number of followers and followees etc. in a generalized linear model, to predict the number of retweets. \newcite{naveed2011bad} extend this approach to perform content-based retweet prediction using several features including sentiments, emoticons, punctuations etc. \newcite{bandari2012pulse} apply the same approach for regression as well as classification, to predict the number of retweets specifically for news articles. \newcite{zaman2014bayesian} present a Bayesian model for retweet prediction using early retweet times, retweets of other tweets, and the user's follower graph. \newcite{tan2014effect} analyze whether different wording of a tweet by the same author affects its popularity. SEISMIC \cite{zhao2015seismic} and PSEISMIC \cite{chen2017pseismic} are statistical methods to predict the final number of retweets. \newcite{zhang2018predicting} approach retweet prediction as a multi-class classification problem, and present a feature-weighted model, where weights are computed using information gain.

\subsection{Training with imbalanced datasets}
\label{sec:imbal}
Various methods to handle imbalanced datasets have been described by \newcite{kotsiantis2006handling}. These include undersampling \cite{kotsiantis2003mixture}, oversampling, and feature selection \cite{zheng2004feature} at the data level. However, due to random undersampling, potentially useful samples can be discarded, while random oversampling poses the risk of overfitting. This problem can be handled at the algorithmic level as well: the threshold method \cite{weiss2004mining} produces several classifiers by varying the threshold of the classifier score. One-class classification can be performed using a divide-and-conquer approach, to iteratively build rules to cover new training instances \cite{cohen1995fast}. Cost-sensitive learning \cite{domingos1999metacost} uses unequal misclassification costs to address the class imbalance problem.

\section{Approaches to authorship verification}
\label{sec:approaches}
As described in Section \ref{sec:methodology}, we build classification models to distinguish between user and non-user tweets. We have explored five distinct approaches to build such models.

\subsection{Approach 1: Compression}
\label{sec:approach1}

This approach is inspired from Kolmogorov complexity \cite{li2013introduction}, which argues that the compressibility of a text reflects the quality of the underlying model.
We use the Lempel-Ziv-Welch (LZW) compression algorithm \cite{welch1984technique} to approximate Kolmogorov complexity by dynamically building a dictionary to encode word patterns from the training corpus. The longest occurring pattern match present in the dictionary is used to encode the text.

We hypothesize that the length of a tweet $T$ from user $U$, compressed using a dictionary built from positive examples, will be less than the length of the same tweet compressed using a dictionary built from negative examples.

We use the following setup to classify test tweets for each Twitter user in our dataset:
\begin{enumerate}
    \item Build an encoding dictionary using positive examples ($\textrm{train}_{\textrm{pos}}$), and an encoding dictionary using negative examples ($\textrm{train}_{\textrm{neg}}$).
    \item Encode the new tweet $T$ using both these dictionaries, to obtain $T_{\textrm{pos}} = \textrm{encode}_{\textrm{pos}}(T)$ and $T_{\textrm{neg}} = \textrm{encode}_{\textrm{neg}}(T)$ respectively.
    \item If the length of $T_{\textrm{pos}}$ is less than that of $T_{\textrm{neg}}$, classify $T$ as positive; else, classify it as negative.
\end{enumerate}

This gives us the class label for each new tweet $T$. In addition, we compute the characterization score of tweet $T$ with respect to user $U$, as described in Equation \ref{eq:compression_score}.

\begin{equation}
\label{eq:compression_score}
\textrm{score}(T, U) = 1 - \frac{\textrm{len}(T_{\textrm{pos}})}{\textrm{len}(T)}
\end{equation}

Thus the shorter the length of the encoded tweet, the more characteristic of the user $T$ is.

\subsection{Approach 2: Topic modeling}

We hypothesize that each user writes about topics with a particular probability distribution, and that each tweet reflects the probability distribution over these topics. We train a topic model using Latent Dirichlet Allocation (LDA) \cite{lda} on a large corpus of tweets, and use this topic model to compute topic distributions for individual tweets. We then use these values as features. We experiment with two types of classifiers: Logistic Regression (LR), and Multi Linear Perceptron (MLP) of size $(5, 5, 5)$.
We represent each tweet as a distribution over $n=500$ topics.

The characterization score of a tweet $T$ is given by the classifier's confidence that $T$ belongs to the positive class.

\subsection{Approach 3: n-gram probability}
\label{sec:approach3}

We hypothesize that a Twitter user can be characterized by usage of words and their frequencies in tweets, and model this using n-gram frequencies.

We use the following setup to classify test tweets for each Twitter user in our dataset:
\begin{enumerate}
    \item Build a frequency dictionary of all n-grams in positive examples ($\textrm{train}_{\textrm{pos}}$), and a frequency dictionary of all n-grams in negative examples ($\textrm{train}_{\textrm{neg}}$).
    \item Compute the average probability of all n-gram sequences in the new tweet $T$ using both these dictionaries, to obtain $\textrm{prob}_{\textrm{pos}}(T)$ and $\textrm{prob}_{\textrm{neg}}(T)$ respectively. Here, we use add-one smoothing and conditional backoff to compute these probability values.
    \item If $\textrm{prob}_{\textrm{pos}}(T)$ is greater than $\textrm{prob}_{\textrm{neg}}(T)$, classify $T$ as positive; else, classify it as negative.
\end{enumerate}

The characterization score of tweet $T$ is given by the average n-gram probability computed using the frequency dictionary of $\textrm{train}_{\textrm{pos}}$. We experiment with $n = 1$ (unigrams) and $n = 2$ (bigrams).

\begin{table*}[!b]
\centering
\resizebox{\textwidth}{!}{%
\begin{tabular}{|c|c|cc|cc|cccc|c|}
\hline
\textbf{User} & \rotatebox[origin=lb]{90}{(1) Compression} & \rotatebox[origin=lb]{90}{(2) LDA + LR} & \rotatebox[origin=lb]{90}{(2) LDA + MLP} & \rotatebox[origin=lb]{90}{(3) Bigram} & \rotatebox[origin=lb]{90}{(3) Unigram} & \rotatebox[origin=lb]{90}{(4) FT + LR} & \rotatebox[origin=lb]{90}{(4) FT + MLP} & \rotatebox[origin=lb]{90}{(4) BERT + LR} & \rotatebox[origin=lb]{90}{(4) BERT + MLP} & \rotatebox[origin=lb]{90}{(5) BERT + LSTM } \\ \hline
Amitabh Bachchan & 72.93 & 69.47 & 74.91 & 84.45 & 90.16 & 84.58 & 87.13 & 93.59 & 93.73 & \textbf{96.32} \\
Ariana Grande & 71.98 & 76.76 & 80.57 & 73.89 & 85.62 & 84.00 & 85.28 & 87.36 & 87.98 & \textbf{90.20} \\
Barack Obama & 78.85 & 80.09 & 87.20 & 82.47 & 92.58 & 91.75 & 93.06 & 95.28 & 95.57 & \textbf{96.57} \\
Bill Gates & 74.29 & 70.78 & 81.78 & 81.21 & 87.10 & 86.34 & 83.97 & 91.56 & 92.41 & \textbf{92.41} \\
Donald Trump & 72.11 & 77.38 & 81.91 & 77.68 & 89.70 & 87.72 & 88.88 & 90.94 & 91.62 & \textbf{93.40} \\
Ellen DeGeneres & 70.75 & 69.39 & 74.53 & 71.90 & 84.40 & 81.27 & 83.11 & 87.38 & 88.60 & \textbf{91.12} \\
J K Rowling & 63.39 & 64.12 & 70.84 & 71.27 & 77.08 & 79.40 & \textbf{80.75} & 79.34 & 80.68 & 79.71 \\
Jimmy Fallon & 67.39 & 72.04 & 73.89 & 78.41 & 85.59 & 82.25 & 83.06 & 85.82 & 86.95 & \textbf{88.62} \\
Justin Bieber & 73.16 & 70.99 & 79.84 & 75.06 & 85.72 & 85.50 & 86.76 & 89.53 & 89.92 & \textbf{92.89} \\
Kevin Durant & 68.78 & 76.56 & 80.24 & 74.61 & 86.21 & 85.76 & 86.93 & 84.75 & 85.84 & \textbf{88.62} \\
Kim Kardashian & 69.73 & 72.10 & 76.39 & 71.49 & 83.89 & 80.92 & 82.49 & 84.12 & 85.25 & \textbf{88.35} \\
Lady Gaga & 66.01 & 67.07 & 72.40 & 71.44 & 81.14 & 76.91 & 79.90 & 81.46 & 83.21 & \textbf{84.54} \\
LeBron James & 67.95 & 66.21 & 73.33 & 74.17 & 82.42 & 81.97 & 77.05 & 83.48 & 84.77 & \textbf{85.53} \\
Narendra Modi & 82.78 & 84.40 & 89.51 & 90.75 & 94.39 & 94.69 & 95.71 & 97.21 & \textbf{97.41} & 97.33 \\
Oprah Winfrey & 68.61 & 61.74 & 70.47 & 75.37 & 83.88 & 83.69 & 83.51 & 86.37 & 87.07 & \textbf{90.01} \\ \hline

\textbf{Mean} & 71.25 & 71.94 & 77.86 & 76.95 & 85.99 & 84.45 & 85.17 & 87.88 & 88.73 & \textbf{90.37} \\ \hline
\end{tabular}%

}
\caption{Test accuracy (\%) of five approaches to classify user vs. non-user tweets (The best performing approach is shown in \textbf{bold} for each user) [Note that for each user, the test set contains an equal number of positive and negative examples.]}
\label{tab:comparison5approaches}
\end{table*}

\subsection{Approach 4: Document embeddings}
\label{sec:approach4}

We hypothesize that if we obtain latent representations of tweets as documents, tweets from the same author will cluster together, and will be differentiable from tweets from others. To that end, we use the following setup:
\begin{enumerate}
    \item We obtain representations of tweets as document embeddings. We experiment with two types of document embeddings: FastText \cite{fasttext} (embedding size = 100) and BERT-Base, uncased \cite{bert} (embedding size = 768).
    \item We then use these embeddings as features to train a classification model. We experiment with two types of classifiers: Logistic Regression (LR) and Multi Linear Perceptron (MLP) of size $(5, 5, 5)$.
\end{enumerate}

The characterization score of tweet $T$ is given by the classifier's confidence that $T$ belongs to the positive class.
\newline
\newline
\noindent
\textbf{Iterative sampling:} As described in Section \ref{sec:methodology}, there exists an extreme class imbalance for this binary classification task, in that the number of negative examples is far more than the number of positive examples. Here, we explore an iterative sampling technique to address this problem. We train our classifier for multiple iterations, coupling the same $\textrm{train}_{\textrm{pos}}$ with a new randomly sampled $\textrm{train}_{\textrm{neg}}$ set in each iteration.

\begin{figure}[H]
  \includegraphics[width=\linewidth]{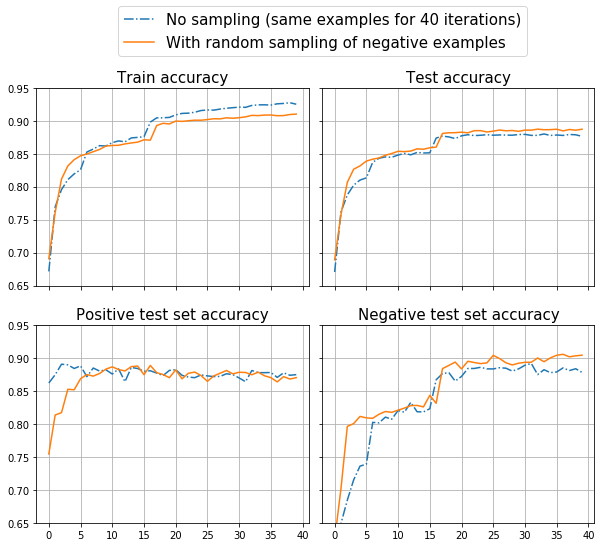}
  \caption{Mean accuracy of the BERT + MLP classifier for all users over 40 iterations}
  \label{fig:iterativesampling}
\end{figure}

We conduct this experiment for all users with the best performing model for this approach, i.e. we use BERT embeddings as features, and MLP for classification. We train this classifier for 40 iterations, and compare the model's performance when we use the same set of negative examples vs. when we randomly sample new negative examples in each iteration.

Figure \ref{fig:iterativesampling} shows the mean train and test accuracy for all users over 40 iterations.
As expected, the training accuracy is higher if we do not sample, as the model gets trained on the same data repeatedly in each iteration. However, if we perform random sampling, the model is exposed to a larger number of negative examples, which results in a higher test accuracy (+ 1.08\%), specifically for negative test examples (+ 2.62\%).

\subsection{Approach 5: Token embeddings and sequential modeling}
\label{sec:approach5}
In this approach, we tokenize each tweet, and obtain embeddings for each token. We then sequentially give these embeddings as input to a classifier.

We use a pretrained model (BERT-Base, Uncased: 12-layer, 768-hidden, 12-heads, 110M parameters) to generate token embeddings of size 768, and pass these to a Long Short Term Memory (LSTM) \cite{lstm} classifier. We use an LSTM layer with 768 units with dropout and recurrent dropout ratio 0.2, followed by a dense layer with sigmoid activation. We train this model using the Adam optimizer \cite{adam} and binary cross-entropy loss, with accuracy as the training metric.

\subsection{Results and Comparison}
Table \ref{tab:comparison5approaches} presents the user-wise test accuracy of the five approaches under the specified configurations. Note that the test set contains an equal number of positive and negative examples for each author.

Other baselines that we attempted to compare against include the best submissions to the PAN 2013 and 2014 author verification challenge: \newcite{pan2013winner} and \newcite{slightlymodifiedimpostor}, which are variants of the Impostors Method. This challenge employed significantly longer documents (with an average of 1039, 845, and 4393 words per document for articles, essays and novels respectively, as opposed to an average of 19 words per tweet) and significantly fewer documents per author (an average of 3.2, 2.6 and 1 document/s per author, as opposed to an average of 6738 tweets per user). Our experiments with the authorship verification classifier \cite{stylo} showed that the Impostors Method is prohibitively expensive on larger corpora, and also performed too inaccurately on short texts to provide a meaningful baseline.

For 13 of the 15 users in our dataset, Approach \ref{sec:approach5} (token embeddings followed by sequential modeling) has the highest accuracy.
This model correctly identifies the author of 90.37\% of all tweets in our study, and will be used to define the characterization score for our subsequent studies.

\section{User study}

To verify whether human evaluators are in agreement with our characterization model, we conducted a user study using MTurk \cite{mturk}.

\subsection{Setup}
\label{sec:setup}
For each user in our dataset, we build a set of 20 tweet pairs, with one tweet each from the 50 top-scoring and bottom-scoring tweets written by the user.
We ask the human evaluator to choose which tweet sounds more characteristic of the user.
To validate that the MTurk worker knows enough about the Twitter user to pick a characteristic tweet, we use a qualification test containing a basic set of questions about the Twitter user.
We were unable to find equal numbers of Turkers familiar with each subject, so our number of evaluators $n$ differs according to author.

\subsection{Results}
\label{sec:mturkresults}

Table \ref{tab:mturk} describes the results obtained in the user study: the mean and standard deviation of percentage of answers in agreement with our model, the p-value, and the number of MTurk workers who completed each task.
We find that the average agreement of human evaluators with our model is \mturkresult \% over all 15 users in our dataset.

\begin{table}[htbp]
\centering
\resizebox{\linewidth}{!}{%
\begin{tabular}{|c|c|r|c|r|}
\hline
\textbf{User} & \textbf{Mean(\%)} & \boldmath$\sigma$\textbf{(\%)} & \textbf{p-value} & \boldmath$n$ \\ \hline
Amitabh Bachchan & 67.08 & 16.44 & \textbf{1.30e-07} & 12 \\
Ariana Grande & 67.19 & 24.01 & \textbf{7.60e-10} & 16 \\
Barack Obama & 55.75 & 17.04 & 2.43e-02 & 20 \\
Bill Gates & 70.26 & 14.19 & \textbf{1.72e-15} & 19 \\
Donald Trump & 83.85 & 8.87 & \textbf{2.52e-58} & 26 \\
Ellen DeGeneres & 73.75 & 14.22 & \textbf{5.44e-22} & 20 \\
J K Rowling & 65.79 & 10.04 & \textbf{7.51e-10} & 19 \\
Jimmy Fallon & 80.00 & 21.93 & \textbf{5.76e-25} & 14 \\
Justin Bieber & 71.94 & 22.57 & \textbf{3.97e-17} & 18 \\
Kevin Durant & 64.38 & 15.04 & \textbf{3.04e-07} & 16 \\
Kim Kardashian & 71.25 & 14.95 & \textbf{9.27e-18} & 20 \\
Lady Gaga & 85.00 & 10.31 & \textbf{1.45e-22} & 9 \\
LeBron James & 63.50 & 12.15 & \textbf{7.38e-08} & 20 \\
Narendra Modi & 60.45 & 13.68 & 2.34e-03 & 11 \\
Oprah Winfrey & 75.79 & 18.12 & \textbf{1.25e-24} & 19 \\ \hline
\end{tabular}%

}
\caption{MTurk user study results: For each of these 15 celebrities, human evaluators support our representativeness scores with a significance level above 0.05. (p-values $<$ $10^{-5}$ are shown in \textbf{bold}.)}
\label{tab:mturk}
\end{table}

For each of the 15 celebrities, the human evaluators agree with our model above a significance level of 0.05, and in 13 of 15 cases above a level of $10^{-5}$.
This makes clear our scores are measuring what we intend to be measuring.

\section{Mapping with popularity}
\label{sec:popularity}

\subsection{Correlation}
\label{sec:correlation}

We now explore the relationship between characterization score and tweet popularity for each of the users in our dataset.
To analyze this relationship, we perform the following procedure for each author $U$:
\begin{enumerate}
    \item Sort all tweets written by $U$ in ascending order of characterization score.
    \item Bucket the sorted tweets by percentile score (1 to 100).
    \item For each bucket, calculate the mean number of likes, replies, and retweets.
    \item Compute the correlation of this mean and the percentile score.
\end{enumerate}

\begin{table}[htbp]
\centering
\resizebox{\linewidth}{!}{%
\begin{tabular}{|r|c|c|c|}
\hline
\multicolumn{1}{|c|}{\textbf{User}} & \textbf{Likes} & \textbf{Replies} & \textbf{Retweets} \\ \hline
Donald Trump & {\color[HTML]{009901} \textbf{0.64}} & {\color[HTML]{009901} \textbf{0.63}} & {\color[HTML]{009901} \textbf{0.55}} \\
Amitabh Bachchan & {\color[HTML]{009901} \textbf{0.58}} & {\color[HTML]{009901} \textbf{0.81}} & {\color[HTML]{009901} \textbf{0.69}} \\
Narendra Modi & {\color[HTML]{009901} \textbf{0.46}} & 0.01 & 0.22 \\
Jimmy Fallon & {\color[HTML]{009901} \textbf{0.29}} & {\color[HTML]{009901} \textbf{0.54}} & {\color[HTML]{009901} \textbf{0.41}} \\
J K Rowling & 0.21 & {\color[HTML]{009901} \textbf{0.32}} & 0.14 \\
Lady Gaga & 0.05 & 0.12 & -0.01 \\
Bill Gates & -0.05 & -0.11 & -0.21 \\
LeBron James & -0.22 & {\color[HTML]{CB0000} \textbf{-0.27}} & -0.24 \\
Oprah Winfrey & {\color[HTML]{CB0000} \textbf{-0.30}} & {\color[HTML]{CB0000} \textbf{-0.41}} & -0.17 \\
Ellen DeGeneres & {\color[HTML]{CB0000} \textbf{-0.34}} & {\color[HTML]{CB0000} \textbf{-0.29}} & {\color[HTML]{CB0000} \textbf{-0.40}} \\
Barack Obama & {\color[HTML]{CB0000} \textbf{-0.45}} & {\color[HTML]{CB0000} \textbf{-0.46}} & {\color[HTML]{CB0000} \textbf{-0.45}} \\
Kevin Durant & {\color[HTML]{CB0000} \textbf{-0.57}} & {\color[HTML]{CB0000} \textbf{-0.67}} & {\color[HTML]{CB0000} \textbf{-0.53}} \\
Kim Kardashian & {\color[HTML]{CB0000} \textbf{-0.71}} & {\color[HTML]{CB0000} \textbf{-0.72}} & {\color[HTML]{CB0000} \textbf{-0.70}} \\
Justin Bieber & {\color[HTML]{CB0000} \textbf{-0.73}} & {\color[HTML]{CB0000} \textbf{-0.50}} & {\color[HTML]{CB0000} \textbf{-0.71}} \\
Ariana Grande & {\color[HTML]{CB0000} \textbf{-0.74}} & {\color[HTML]{CB0000} \textbf{-0.77}} & {\color[HTML]{CB0000} \textbf{-0.75}} \\ \hline
\end{tabular}%
}
\caption{Pearson correlation coefficients between mean popularity measure and percentile, for each user (Coefficients with p-value $<$ 0.01 are shown in \textbf{bold} color). {\color[HTML]{009901}{Green}} values exhibit significant positive correlation, and {\color[HTML]{CB0000}{red}} values significant negative correlation.}
\label{tab:pearson}
\end{table}

The Pearson correlation coefficients (r-values) are listed in Table \ref{tab:pearson}. The users at the top (Trump, Bachchan, Modi) all display very strong positive correlation. We name this group UPC (Users with Positive Correlation), and the group of users at the bottom (Grande, Bieber, Kardashian) as UNC (Users with Negative Correlation).

\subsection{Interpretation}
\label{sec:interpretation}
For users with positive correlation, the higher the tweet's characterization score, the more popular it becomes, i.e. the more likes, replies, and retweets it receives. In contrast, for users with negative correlation, the higher the tweet score, the less popular it becomes.

\begin{figure}[htbp]
  \includegraphics[width=\linewidth]{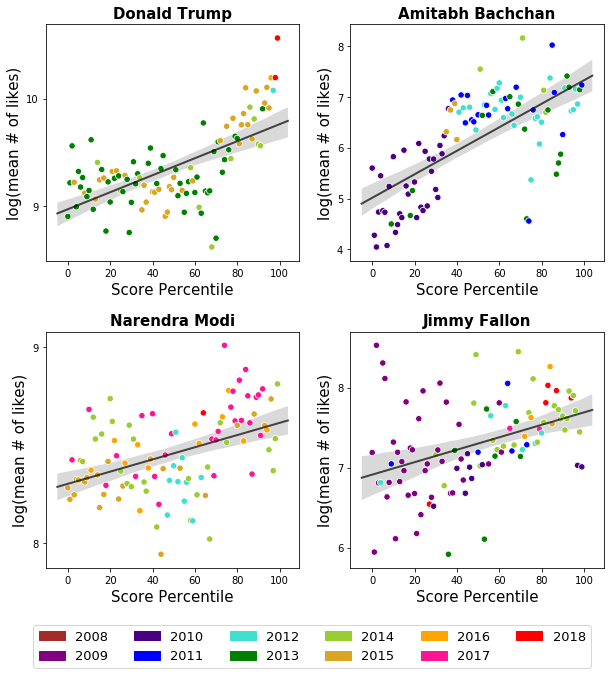}
  \caption{Log mean likes vs. percentile for users of positive correlation (The color denotes the year for which maximum tweets are present in the percentile bucket).}
  \label{fig:upc}
\end{figure}

\begin{figure}[H]
  \includegraphics[width=\linewidth]{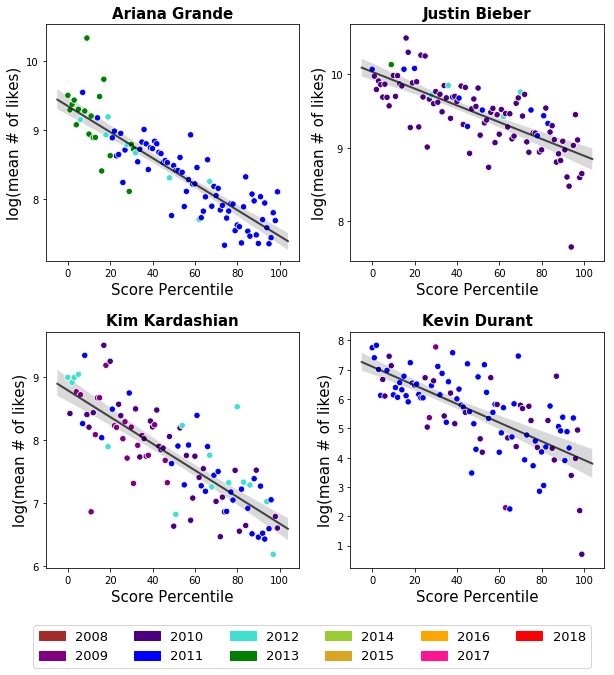}
  \caption{Log mean likes vs. percentile for users of negative correlation (The color denotes the year for which maximum tweets are present in the percentile bucket).}
  \label{fig:unc}
\end{figure}

Figure \ref{fig:upc} shows the plot of log mean number of likes per bucket vs. tweet score percentile, for users with the highest positive correlation.
Similarly, Figure \ref{fig:unc} shows the plot of log mean number of likes per bucket vs. tweet score percentile, for users with the highest negative correlation.

One may question whether these results are due to temporal effects: user's popularity vary with time, and perhaps the model's more characteristic tweets simply reflect periods of authorship.
Figures \ref{fig:upc} and \ref{fig:unc} disprove this hypothesis.
Here the color of each point denotes the year for which most tweets are present in the corresponding bucket. Since the distribution of colors over time is not clustered, we infer that the observed result is not an artifact of temporal effects. In both cases, there is a strong trend in tweet popularity based on tweet score. We note that the plots are presented on the log scale, meaning the trends here are exponential.

\subsection{Qualitative Analysis}

We present examples of the {\color[HTML]{009901}most} and {\color[HTML]{CB0000}least} characteristic tweets for celebrities from three categories, along with their corresponding characterization scores computed using Approach \ref{sec:approach5}.

\subsubsection{Users with Positive Correlation (UPC)}

\begin{table}[H]
\caption*{\textbf{Donald Trump}}
\resizebox{\linewidth}{!}{%
\begin{tabular}{|p{8cm}|l|}
\hline
\multicolumn{1}{|c|}{\textbf{Tweet}} & \multicolumn{1}{c|}{\textbf{Score}} \\ \hline
\noindent{\color[HTML]{009901} Prior to the election it was well known that I have interests in properties all over the world. Only the crooked media makes this a big deal!} & {\color[HTML]{009901} 0.9998}       \\ \hline
\noindent{\color[HTML]{CB0000} Today is the first day of the rest of your life - make the most of it!}    & {\color[HTML]{CB0000} 0.0001}       \\ \hline
\end{tabular}
}
\label{tab:trump_upc}
\end{table}
\vspace{-5mm}
\begin{table}[H]
\caption*{\textbf{Amitabh Bachchan}}
\resizebox{\linewidth}{!}{%
\begin{tabular}{|p{8cm}|l|}
\hline
\multicolumn{1}{|c|}{\textbf{Tweet}} & \multicolumn{1}{c|}{\textbf{Score}} \\ \hline
\noindent{\color[HTML]{009901} T 2843 - The work is demanding .. the crew binding .. the city exciting .. and the dialogues expanding ..
`BADLA' is grinding .. !!} & {\color[HTML]{009901} 0.9996}       \\ \hline
\noindent{\color[HTML]{CB0000} hahaha .. now i dont have a HD .. but ya a car ride is on ..}    & {\color[HTML]{CB0000} 0.0002}       \\ \hline
\end{tabular}
}
\label{tab:bachchan_upc}
\end{table}

The characterization score appears to have correctly captured aspects of the user's personality from their corpus of tweets. For these celebrities, high scoring tweets generally prove more popular (In this example - Donald Trump: {\color[HTML]{009901}70.5K} vs. {\color[HTML]{CB0000}693} likes; Amitabh Bachchan: {\color[HTML]{009901}7.1K} vs. {\color[HTML]{CB0000}9} likes), as reflected in their positive correlation coefficients.

\subsubsection{Users with Negative Correlation (UNC)}

\begin{table}[H]
\caption*{\textbf{Ariana Grande}}
\resizebox{\linewidth}{!}{%
\begin{tabular}{|p{8cm}|l|}
\hline
\multicolumn{1}{|c|}{\textbf{Tweet}} & \multicolumn{1}{c|}{\textbf{Score}} \\ \hline
\noindent{\color[HTML]{009901} Finalizing the set list for Fresno! Getting so excited.. Can't believe the show is already almost sold out, you guys are amazing. Xoxo!} & {\color[HTML]{009901} 0.9997}       \\ \hline
\noindent{\color[HTML]{CB0000} The first thing I do when I get to a new city is look up how close the nearest Whole Foods is.}    & {\color[HTML]{CB0000} 0.0002}       \\ \hline
\end{tabular}
}
\label{tab:grande_unc}
\end{table}
\begin{table}[H]
\caption*{\textbf{Justin Bieber}}
\resizebox{\linewidth}{!}{%
\begin{tabular}{|p{8cm}|l|}
\hline
\multicolumn{1}{|c|}{\textbf{Tweet}} & \multicolumn{1}{c|}{\textbf{Score}} \\ \hline
\noindent{\color[HTML]{009901} grateful to everyone who came out and to my band, dancers, and whole crew.  The energy last night was incredible and cant wait to tour} & {\color[HTML]{009901} 0.9999}       \\ \hline
\noindent{\color[HTML]{CB0000} Less cantaloupe, more berries. I'm talking to you, pre-packaged fruit salads. Don't play me like that.}    & {\color[HTML]{CB0000} 0.00002}       \\ \hline
\end{tabular}
}
\label{tab:bieber_unc}
\end{table}

Again, high scoring tweets appear more characteristic of their respective users. But here, low scoring tweets are generally more popular (In this example - Ariana Grande: {\color[HTML]{009901}622} vs. {\color[HTML]{CB0000}2.4K} likes; Justin Bieber: {\color[HTML]{009901}454} vs. {\color[HTML]{CB0000}13.8K} likes), as reflected in their negative correlation coefficients.

\subsubsection{Users with no significant correlation}

\begin{table}[H]
\caption*{\textbf{Bill Gates}}
\resizebox{\linewidth}{!}{%
\begin{tabular}{|p{8cm}|l|}
\hline
\multicolumn{1}{|c|}{\textbf{Tweet}} & \multicolumn{1}{c|}{\textbf{Score}} \\ \hline
\noindent{\color[HTML]{009901} I recently visited a lab doing super-cool energy work-a good reminder of why governments should sponsor R\&D} & {\color[HTML]{009901} 0.9986}       \\ \hline
\noindent{\color[HTML]{CB0000} There's a lot of green on this map-which is good-but still not enough.}    & {\color[HTML]{CB0000} 0.0027}       \\ \hline
\end{tabular}
}
\label{tab:bill_gates}
\end{table}

Here, tweets from extreme ends of the spectrum have similar content, so little variation can be expected in their popularity. For this celebrity, there is no significant correlation between characterization score and popularity.

\section{Conclusions}
We have presented and evaluated measures of binary author classification, to obtain a user-specific characterization score for each tweet. We demonstrate that sequential modeling on word embeddings yields the best result of 90.37\% mean test accuracy, and that human evaluators are in agreement with our model \mturkresult\% of the time.
Our work demonstrates that representativeness scores correlate with popularity, and opens new research directions concerning virality on social media.

\section*{Acknowledgments}
We are grateful to the anonymous reviewers for their helpful feedback. We also thank Niranjan Balasubramanian and H. Andrew Schwartz for their comments and suggestions. This work was partially supported by NSF grants IIS-1546113 and IIS-1927227.

\bibliography{emnlp-ijcnlp-2019}
\bibliographystyle{acl_natbib}

\end{document}